\PassOptionsToPackage{table}{xcolor}
\documentclass{article} % For LaTeX2e
\usepackage{iclr2026_conference,times}

\usepackage{amsmath,amsfonts,bm}

\def\eqref#1{equation~\ref{#1}}
\def\Eqref#1{Equation~\ref{#1}}
\def\1{\bm{1}}

\DeclareMathAlphabet{\mathsfit}{\encodingdefault}{\sfdefault}{m}{sl}
\SetMathAlphabet{\mathsfit}{bold}{\encodingdefault}{\sfdefault}{bx}{n}

\def\gE{{\mathcal{E}}}

\def\gG{{\mathcal{G}}}

\def\gV{{\mathcal{V}}}

\usepackage{url}

\usepackage[T1]{fontenc}
\usepackage{amsmath,amsfonts,amssymb,amsthm,mathtools,bm}
\usepackage{booktabs}
\usepackage{nicefrac}
\usepackage{microtype}
\usepackage{xcolor}
\usepackage{enumitem}
\usepackage{multirow}
\usepackage{array}
\usepackage{color}
\usepackage[table]{xcolor} 
\usepackage{wrapfig}
\usepackage{multicol}
\usepackage{caption}
\usepackage{diagbox}
\usepackage{pifont}
\usepackage{pdfpages}
\usepackage{arydshln}
\usepackage{makecell}
\usepackage{graphicx}
\usepackage{adjustbox}
\usepackage{colortbl}

\definecolor{blue(pigment)}{rgb}{0.2, 0.2, 0.6}
\definecolor{lightgray}{rgb}{0.88, 0.88, 0.88}
\definecolor{blanchedalmond}{rgb}{1.0, 0.9, 0.8}
\definecolor{tearose}{rgb}{0.96, 0.76, 0.76}
\definecolor{languidlavender}{rgb}{0.84, 0.79, 0.9}
\definecolor{lavender(web)}{rgb}{0.99, 0.93, 0.93}

\newcommand{\hj}[1]{\textcolor{black}{#1}}
\newcommand{\sh}[1]{\textcolor{black}{#1}}
\usepackage[colorlinks=true, linkcolor=blue(pigment), citecolor=blue(pigment)]{hyperref}
\usepackage[capitalise,noabbrev,nameinlink]{cleveref}

\crefformat{figure}{Figure~#2{\color{blue(pigment)}#1}#3}
\crefformat{table}{Table~#2{\color{blue(pigment)}#1}#3}
\crefformat{equation}{Equation~#2{\color{blue(pigment)}#1}#3}
\crefformat{section}{Section~#2{\color{blue(pigment)}#1}#3}
\crefformat{appendix}{Appendix~#2{\color{blue(pigment)}#1}#3}

\def\pz{{\phantom{0}}}

\title{Learning Flexible Forward Trajectories for \\Masked Molecular Diffusion}

\author{
  Hyunjin Seo\textsuperscript{1,2}$^*$, 
  Taewon Kim\textsuperscript{1,2}$^*$, 
  Sihyun Yu\textsuperscript{1},
  SungSoo Ahn\textsuperscript{1}\\
  Korea Advanced Institute of Science and Technology (KAIST)\textsuperscript{1}, Polymerize\textsuperscript{2}\\
  \texttt{\{bella72, maxkim139, sihyun.yu, sungsoo.ahn\}@kaist.ac.kr}
 }

\iclrfinalcopy % Uncomment for camera-ready version, but NOT for submission.
\begin{document}

\maketitle

\begin{abstract}
Masked diffusion models (MDMs) have achieved notable progress in modeling discrete data, while their potential in molecular generation remains underexplored. In this work, we explore their potential and introduce the surprising result that na\"ively applying standards MDMs to molecules leads to severe performance degradation. We trace this critical issue to a \emph{state-clashing problem}---where the forward diffusion trajectories of distinct molecules collapse into a common state, resulting in a mixture of reconstruction targets that cannot be learned with a typical reverse diffusion with unimodal predictions. To mitigate this, we propose \textbf{M}asked \textbf{E}lement-wise \textbf{L}earnable \textbf{D}iffusion (\system{}) that orchestrates per-element corruption trajectories to avoid collisions between different molecular graphs. This is realized through a parameterized noise scheduling network that learns distinct corruption rates for individual graph elements, \ie, atoms and bonds. Across extensive experiments, \system{} is the first diffusion-based molecular generator to achieve 100\% chemical validity in unconditional generation on QM9 and ZINC250K datasets, while markedly improving distributional and property alignment over standard MDMs.
\end{abstract}

\renewcommand{\thefootnote}{\fnsymbol{footnote}}
\addtocounter{footnote}{0} % reset
\footnotetext[1]{Equal Contribution.}
\footnotetext[2]{Corresponding Author.}
\section{Introduction}\label{1_introduction}
Molecular generation is critical in a variety of real-world applications, such as drug discovery~\citep{drug_discovery} and material design~\citep{mat_design, mat_diffusion}. However, the task remains challenging due to the extremely large and complex nature of the chemical space~\citep{du2024machine}. With the remarkable recent progress in deep generative models~\citep{vae, normalizing_flow, d3pm, llmsurvey}, many approaches have attempted to tackle this problem by training a neural network that learns molecular distributions from large molecular datasets, demonstrating a strong promise in accelerating molecule discovery~\citep{graphga, jtvae, graphaf, gdss, digress, defog}.

In particular, recent works have focused on exploring generative models based on denoising diffusion or flow-matching models,~\citep{gdss, mood, digress, grapharm, grum, graphdit}, to learn a molecular distribution, inspired by their great success in other data domains with scabaility~\citep{ddpm, sde, d3pm, improved_ddpm, sit, vdm, mulan, wan2025}. These models learn to recover original molecules from corrupted versions through a denoising process, where the corruption typically involves altering types of atoms and bonds (\eg, changing a carbon atom to nitrogen, or a single bond to a double bond).

% ~\citep[MDMs;][]{d3pm, maskgit, md4, subs}
Meanwhile, researchers have explored masked diffusion models (MDMs; \citealt{d3pm, maskgit, md4, subs}). Unlike conventional diffusion models that typically design diffusion processes in continuous space, MDMs are specialized for discrete data by defining a diffusion process more suitable in discrete space. Specifically, MDMs define the forward process as element masking and train the model to infill the masked element during the reverse process. Intriguingly, MDMs show great stability and scalability, being comparable or even better than previous generative models for discrete data, such as autoregressive language models~\citep{autoreg1, autoreg2} or high-resolution text-to-image generation~\citep{chang2023muse}. Despite their success in other domains, applying MDMs to molecular graphs is still underexplored.

\begin{figure}[t!]
    \vskip -5pt
    \centering
    \includegraphics[width=\linewidth]{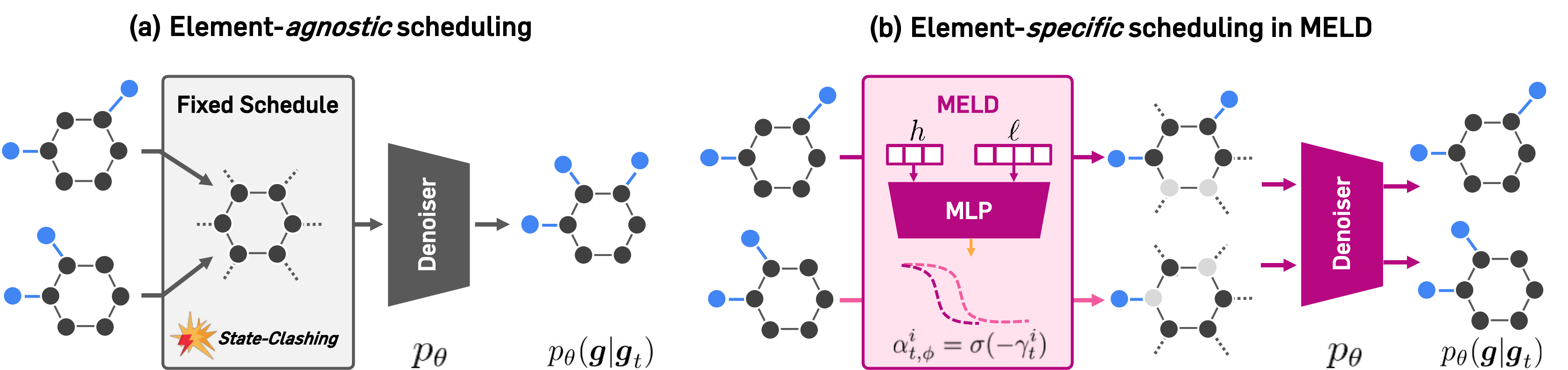}
    \caption{
    Comparison between (a) element-agnostic noise scheduling and (b) element-specific noise scheduling. 
    The former results in an issue denoted as \emph{state-clashing}, leading to generation of invalid molecules. \system{} mitigates this with element-specific noise schedule, effectively orchestrating the forward process to minimize state-clashings.
    }
    \label{fig:framework}
    \vskip -10pt
\end{figure}

In this work, we focus on applying MDMs for molecular generation. Surprisingly, unlike other domains, a naïve adaptation of existing MDMs to molecular graphs results in significantly worse performance than other recent molecular generation methods, often generating distributionally misaligned structures. We argue that this phenomenon stems from \emph{state-clashing problem}: Molecular graphs with different properties and semantics easily collapse into a common intermediate state in the forward process (see~\cref{fig:framework}(a) for an illustration).  
We attribute this to the noise scheduling design in existing MDMs, which assumes a fixed, uniform transition probability across all nodes and edges. However, MDMs learn the reverse process as unimodal predictions, which prevents them from capturing highly multimodal distribution of the true posterior. As a result, they often generate samples that deviate from the target distribution \hj{and, in some cases, violate the chemical rules}. 
%We further describe \emph{state-clashing problem} in~\cref{subsec:clashing}

% Thus, resulting in chemically invalid samples.

% This leads to generating samples that violate the chemical rules or significantly deviate from the target distribution.

To address this, we introduce \system{} (\textbf{M}asked \textbf{E}lement-wise \textbf{L}earnable \textbf{D}iffusion), a novel MDM for molecular graph generation. The main idea of our method is to alleviate the state-clashing problem by proposing an \emph{element-wise learnable forward process}, which generates corruption trajectories in the way of minimizing the occurrence of potential collision. To this end, we introduce a parameterized noise scheduling network to yield distinct corruption rates for individual graph elements (\ie, for nodes or edges). During training, we jointly optimize the forward (\ie, noise scheduling network) and the reverse process (\ie, MDM denoiser network). Intuitively, by assigning per-element trajectories, \system{} organizes the forward process such that the probability of molecules being collapsed to the same intermediate state (see~\cref{fig:framework}(b)) is minimized. Through such evasion, ~\system{} \hj{effectively learns to produce samples capturing the target molecular distribution}.

We evaluate \system{} on diverse molecular datasets, including QM9~\citep{qm9}, Polymers~\citep{polymer}, ZINC250K~\citep{zinc}, Guacamol~\citep{guacamol}, and a synthetic graph benchmark~\citep{sbm}. \hj{First, we demonstrate that \system{} yields substantial improvements in distributional similarity over standard MDMs. Moreover, it is the first molecular diffusion model to achieve 100\% chemical validity on both QM9 and ZINC250K unconditional benchmarks. In conditional generation, \system{} further enhances property alignment by up to 13.4\% over state-of-the-art baseline.} \sh{Finally, we show the scabability and generalzability of \system{} in large-sacle molecule datasets and non-molecule graph datasets, using Guacamol \citep{guacamol} and SBM \citep{sbm}, respectively.}
% Furthermore, compared with other state-of-the-art molecular generation or optimization methods, \system{} achieves comparable or superior performance in both  unconditional and property-conditioned molecule generation tasks. For instance, on the conditional generation setup on the polymer dataset, \system{} achieves up to a 16.5\% enhancement in property alignment compared to the state-of-the-art diffusion model.
% xx.xx measured with XX metric, which is xx.x\% better than the previous state-of-the-art of xx.xx.  

Our contributions are threefold:
\begin{itemize}[topsep=0pt,itemsep=1mm, parsep=0pt, leftmargin=5mm]
    \item We identify a key limitation in applying standard masked diffusion models to molecular generation, the use of an element-agnostic noise schedule, which leads to frequent \textit{state-clashing}.
    \item We present \system{}, a novel masked diffusion framework that mitigates the state-clashing problem by learning per-element noise schedules, allowing adaptive corruption trajectories tailored to individual molecular components.
    \item \system{} significantly improves the overall quality of generated molecules over standard MDM baselines, and surpasses existing molecular \hj{diffusion models in both unconditional and property-conditioned generation tasks. Moreover, its efficacy generalizes consistently to large-scale molecule and synthetic graph benchmarks.}
\end{itemize}

\section{Related work}\label{5_related_works}
\paragraph{Masked diffusion models (MDMs).}
MDMs have emerged as a powerful generative modeling scheme for discrete data generation. Initially, D3PM~\citep{d3pm} introduces an absorbing mask token into the forward process and establishes a conceptual bridge between discrete diffusion and masked language modeling. Additionally, in image generation, MaskGIT~\citep{maskgit} shows that generative modeling based on unmasking enables fast and qualitatively comparable high-fidelity image synthesis compared with left-to-right autoregressive decoding.
More recent efforts have further refined MDMs to close the performance gap with autoregressive models (AR;~\citealt{attention, autoreg1, autoreg2}). Notably, MD4~\citep{md4} and MDLM~\citep{subs} show that the diffusion objective can be simplified as a weighted integral of cross-entropy and that the model can achieve state-of-the-art results over prior diffusion models. Our work also focuses on MDMs, but we explore applying MDMs to molecules.
% Overall, these approaches depart from conventional AR by discarding fixed generation order and instead denoising all elements jointly through iterative, parallel refinement.

\paragraph{Diffusion models for molecules.}
The success of diffusion models for image~\citep{stable_diffusion} and text generation~\citep{diffusion-lm} has inspired researchers to explore diffusion models in molecule domain. A surge of studies~\citep{digress, gdss, mars, grapharm, grum, graphdit} have been proposed to generate de novo molecules, competing with sequential models~\citep{segler2018generating, jtvae, graphaf, jang2024graph, jang2024a} that iteratively constructs a graph by adding graph elements. These efforts can be categorized into two approaches: (1) Score-based molecule diffusion approaches~\citep{gdss, mood, grum} adopt continuous noise on molecular graphs using stochastic differential equations (SDEs)~\citep{sde}. They train a score function to approximate reverse SDEs, relaxing discrete atoms/bonds into a continuous space. 
(2) Discrete diffusion-based approaches~\citep{digress, graphdit, mudiff, trainingfree_discrete} apply discrete noise through Markovian transitions to nodes and edges in molecular graphs. Then they train a denoising neural network to reconstruct perturbed atom and bond types. 

Despite progress in these two directions, masked diffusion frameworks remain underexplored for molecular generation. A preliminary application was explored in \cite{grapharm}, but it generates atoms in an autoregressive manner, limiting its ability to exploit the parallelized reconstruction of MDMs. In contrast, we propose MDMs for molecular graphs by focusing on the state-clashing problem occurring in the forward process, while preserving the parallelism inherent to MDMs.
% Moreover, our method can reconstruct multiple tokens in parallel while achieving the better generation quality.

% Beyond unconditional setting, several recent works~\citep{digress, mood, graphdit} address the property-conditioned generation task by utilizing techniques in vision domains~\citep{classifier_guidance, cfg}. Specifically, DiGress~\citep{digress} and MOOD~\citep{mood} employ an auxiliary property predictor to adjust reverse diffusion probabilities using predictor gradients, guiding the generation of molecules with desired traits. GraphDiT~\citep{graphdit} integrates classifier-free guidance technique by training the denoiser on a mixture of property-conditioned and unconditioned inputs, and interpolates between them during the generation phase. However, all these approaches rely on unified noise scheduling across all molecular constituents, overlooking the risk of state-clashing problem.

\section{\system{}: Masked element-wise learnable diffusion}\label{3_proposed_method}
In this section, we introduce \system{}, a masked diffusion model (MDM) for molecular graph generation that jointly learns per-graph-element corruption rate and the denoising model. As we will explain, our proposed design alleviates the state-clashing problem by producing easily distinguishable forward trajectories for each molecular component. 

\subsection{Main algorithm}
Similar to standard diffusion models~\citep{ddpm, sit, improved_ddpm, dit}, we characterize our masked molecular diffusion process using a forward and reverse process. The forward process progressively corrupts a clean molecular graph through a sequence of Markov transitions, while the reverse process starts from a sample from the prior distribution and iteratively reconstructs the original sample. Unlike the existing work on molecular diffusion models, our key difference is to (1) use masking/unmasking-based forward/reverse transitions (respectively), (2) introduce element-wise parameterization of the forward process, and (3) jointly learn the forward process in addition to the reverse process.

\paragraph{Problem definition.}
We consider the task of generating molecules represented as graphs, denoted by $\gG=(\gV,\gE)$, where $\gV$ is a set of $N$ atoms (nodes) and $\gE$ a set of bonds (edges). We encode $\gV$ with a categorical feature $\bm{x} = (x^{i})_{i=1}^N$, where each $x^i \in \{1,\dots,A\}$ indicates the atom type of node $i$ among the $A>0$ possible types with one category reserved for $[\texttt{mask}]$ token. Similarly, we encode $\gE$ as a categorical feature $\bm{e} = (e^{ij})_{i,j=1}^N$ where each $e^{ij} \in \{1,\dots,B\}$ is the bond type between the atoms $i$ and $j$, with two categories reserved for the ``no bond'' and $[\texttt{mask}]$ tokens.  
%and we aim to train a model to generate $\bm{x}$ and $\bm{e}$. 
%In the remainder of this paper, we denote random variables taking values in $x^{i}$ and $e^{ij}$ as $X^{i}$ and $E^{ij}$, respectively. 
Since the graph is an abstract object, we rely on its explicit representation $\bm{g} = (\bm{x},\bm{e})$ of graph $\mathcal{G}$. %Although this is not unique due to permutation, we note how prior works~\citep{xx} verified that equivariance is not strictly necessary to achieve good performance for generative models. 
Finally, our goal is to train a graph generator from the data distribution $q(\bm{g})$. To this end, we introduce a forward process $q_{\phi}(\bm{g}_{t} | \bm{g}_{t-1})$ and a reverse process $p_{\theta}(\bm{g}_{t-1}|\bm{g}_{t})$, parameterized by $\phi$ and $\theta$, respectively. %Again, we emphasize that our work primarily differs from prior works~\citep{digress, gdss, mood, grum, graphdit} due to the trainable forward diffusion process with an element-wise parameterization, which is crucial for alleviating the state-clashing issue, in which we explain~\cref{subsec:clashing}.
%Further more, the diffusion processes operates in a discrete space augmented with an additional $[\texttt{mask}]$ token. Specifically, we append an additional mask state into original $A$ node categories and $B$ edge categories as $(A+1)$-th and $(B+1)$-th category, respectively.

\paragraph{Forward and reverse diffusion processes.} 
At each timestep $t$, the forward diffusion process $q_{\phi}$ progressively corrupts atoms and bonds by stochastically transitioning them into a masked state, denoted by $\texttt{[mask]}$. This transition is governed by Markov kernels defined with a monotonically increasing sequence $\beta_t\in[0,1]$. Specifically, the Markov kernels are defined as
\begin{equation}
    q_{\phi}(x_{t}^{i}\,|\,x^i_{t-1}) = 
    \begin{cases}
        \beta_{t, \phi}^{i} &\text{ if } x_{t}^{i} = [\texttt{mask}]\\
        1-\beta_{t, \phi}^{i} &\text{ if }x_{t}^{i}=x_{t-1}^{i}
    \end{cases},
    ~
    q_{\phi}(e_{t}^{ij}\,|\,e^{ij}_{t-1}) = 
    \begin{cases}
        \beta_{t, \phi}^{ij} &\text{ if } e_{t}^{ij} = [\texttt{mask}]\\
        1-\beta_{t, \phi}^{ij} &\text{ if }e_{t}^{i}=e_{t-1}^{ij}
    \end{cases}.
\label{eq:cat_sampling}
\end{equation}
Note that prior work~\citep{digress, gdss, mood, grum, graphdit} has assumed the forward diffusion process as a \emph{fixed} transition probability that is uniform across elements (vertices and edges). In contrast, our work introduces an additional degree of freedom, represented by the parameter $\phi$, which allows flexible modeling of the forward diffusion process to overcome the state-clashing problem. Importantly, the kernels are distinct for each vertex, which means that the forward diffusion process is no longer equivariant, \ie, the marginal distribution of an intermediate state is affected by permutation.

Next, the reverse process recovers the clean graph from a noisy state $\bm{g}_t$, and we learn this process through a model $p_\theta(\bm{g}_{t-1}|\bm{g}_t)$. In particular, each denoising step is formalized as the product over nodes and edges~\citep{digress}:
\begin{equation}\label{eq:factorization}
    p_\theta(\bm{g}_{t-1}|\bm{g}_t)=\prod_{v\in\gV}p_\theta(x_{t-1}|\bm{g}_t)\prod_{e \in \gE}p_\theta(e_{t-1}|\bm{g}_t).
\end{equation}
%Note that we parameterize the reverse process using an equivariant neural network, which .

\paragraph{Training objective.}
%In practice, 
The denoising network is trained to directly predict the original graph $\bm{g}_0$ from the noisy intermediate state $\bm{g}_t$, eliminating the need for recursive sampling~\citep{digress, graphdit}. 
This is achieved by training the model with a cross-entropy loss between the predicted logits and the ground-truth node and edge categories~\citep{md4, subs}:
\begin{equation}\label{eq:diffusion}
    \mathcal{L} (\theta, \phi) :=-\mathbb{E}_{t, \bm{g}, \bm{g}_{t}}\bigg[ \sum_{1 \leq i \leq N} \frac{{\dot{\alpha}_{t,\phi}^{i}}}{1-\alpha_{t, \phi}^{i}}\log p_\theta(x^i \mid \bm{g}_t) + \lambda \sum_{1\leq i < j \leq N} \frac{{\dot{\alpha}_{t, \phi}^{ij}}}{1-\alpha_{t, \phi}^{ij}}\log p_\theta(e^{ij} \mid \bm{g}_t)\bigg],
\end{equation}
where the expectation is over $\bm{g}\sim q(\cdot), \bm{g}_{t} \sim q_{\phi}(\cdot|\bm{g})$, $t\sim \operatorname{Unif}[0,1]$. Furthermore, we let $\alpha_{t, \phi}^{i}=\prod_{t'=t}^{T}\beta_{t',\phi}^{i}$ and ${\dot{\alpha}_{t,\phi}^{i}}$ is the derivative $\alpha_{t,\phi}^{i}$ with respect to $t$, and $\lambda>0$ is a hyperparameter that balances the contributions of node- and edge-level reconstruction, following prior works~\citep{digress, graphdit}. Note that na\"ively applying gradient-based optimization to Monte Carlo estimation of the loss (\Eqref{eq:diffusion}) does not incorporate the dependency on $\phi$. We use straight-through gumbel softmax trick (STGS; \citealt{stgs}) to address this, as explained further in~\cref{subsec:optim}.

\subsection{Formalizing the state-clashing problem}\label{subsec:clashing}
\begin{figure}[t!]
    \centering
    \includegraphics[width=\linewidth]{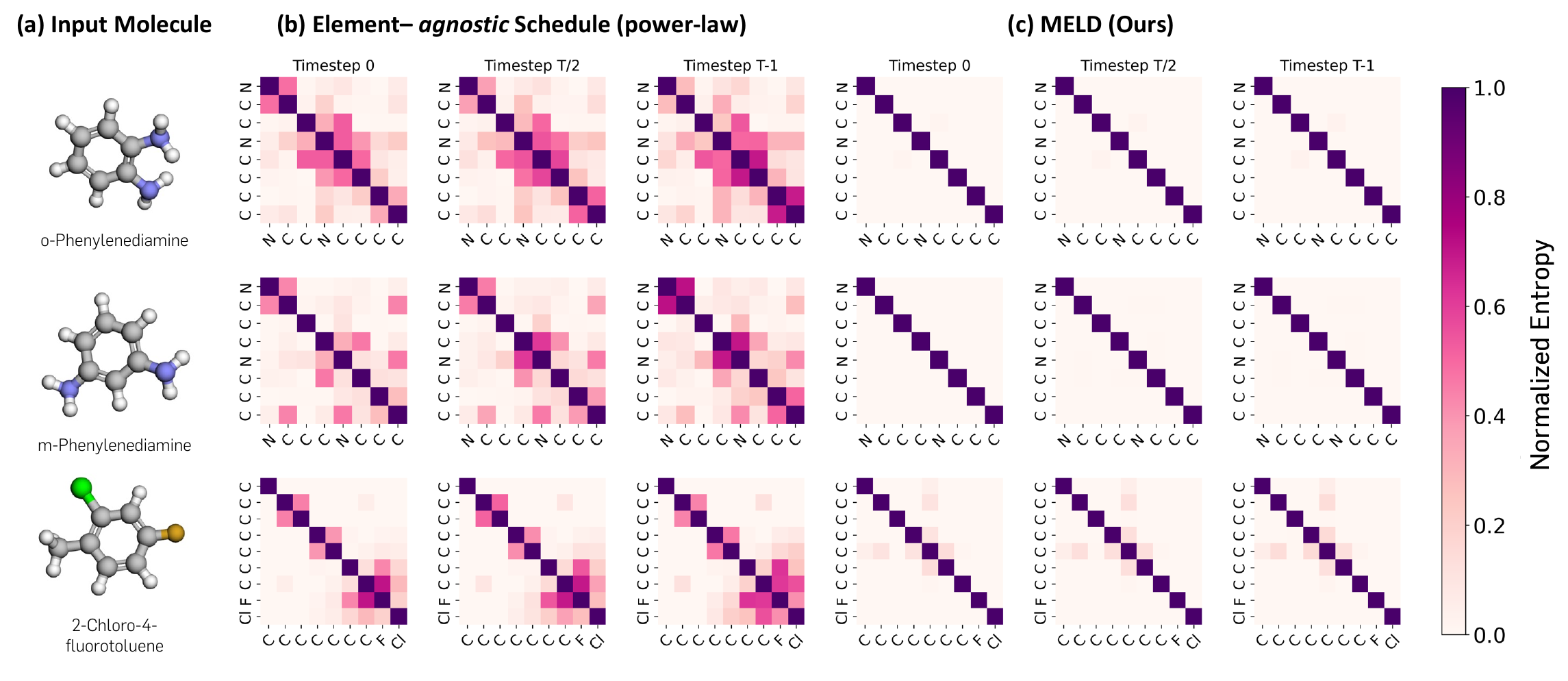}
    \vskip -5pt
    \caption{
    Visualization of prediction entropy for various molecule types. The first and second rows show prediction matrices with nitrogen bonds masked, while the third row shows generations with chlorine bond masked. From left to right: (a) 3D renderings of the input molecules, (b) predictions from MDMs using a fixed power law noise schedule, and (c) predictions from \system{}. Brighter colors indicate lower uncertainty (\ie, higher confidence). \hj{The dark diagonal entries reflect enforced uniform predictions, as self-connections in molecules are not meaningful and are excluded from valid outputs.} Note that predictions are being made for all locations, regardless of their entropy values.
    }
    \label{fig:obs_heatmap}
    \vskip -10pt
\end{figure}

Here, we describe the \emph{state-clashing problem} which naturally arise for training MDMs on graphs without learning the forward process, \ie, set $\beta_{t,\phi}^{i}$ to some constant $\beta_{t}$ for all node $i$ and similarly for the edges. In a nutshell, state-clashing refer to the phenomenon where semantically distinct molecules are corrupted into the same intermediate state, due to the nature of the constant forward process in MDMs. Consequently, the model trained with such constant forward process struggles to infer the correct reconstruction target, resulting in outputs that fail to preserve structural or molecular coherence with target distribution (see~\cref{fig:framework} for an illustration). This is particularly pronounced in molecules with symmetric motifs, to which the number of immediate parent states grows by the number of permutations that leave the motif invariant.

Formally, note that the diffusion model loss in~\cref{eq:diffusion} can be expressed as:
\begin{equation}
    \mathbb{E}_{t}\big[\operatorname{KL}(p(\bm{g}|\bm{g}_{t}), p_{\theta}(\bm{g}|\bm{g}_{t}))\big], \quad p(\bm{g}|\bm{g}_{t}) \propto p(\bm{g}_{t}|\bm{g})p(\bm{g}).
\end{equation}
The main problem is that $p(\bm{g}|\bm{g}_{t})$ can be highly \emph{multimodal}, \ie, there exists many graph $\bm{g}$ with non-zero probabilities of $p(\bm{g}_{t}|\bm{g})$. However, 
the parameterized diffusion model $p_{\theta}(\bm{g} | \bm{g}_{t}) = \prod_{1\leq i \leq N}p(x^{i} | \bm{g}_{t})\prod_{1\leq i<j \leq N}p(e^{ij} | \bm{g}_{t})$ is \emph{unimodal}, as it predicts each node and edge independently, typically resulting in a single mode centered around an average graph.
Furthermore, due to the mode-covering property of KL divergence, the reverse diffusion model trained with~\cref{eq:diffusion} tends to converge to a high-entropy distribution--the model compensates for its inability to represent multiple modes by spreading its probability mass broadly around the single mode, as illustrated in the first and second rows of~\cref{fig:obs_heatmap}. %\footnote{Following prior works~\citep{gdss, grum, graphdit}, we exclude implicit hydrogen atoms.} 
In addition, we visualize the model’s predictions for 2-Chloro-4-fluorotoluene when only the Chlorine bond is masked (\ie, hidden). Owing to the inherent asymmetry of the pyrimidine backbone, the state-clashing issue is less pertinent than Phenylenediamine isomers. As a result, we observe increased prediction confidence even in MDMs employing unified scheduling, underscoring the necessity of addressing the state-clashing.

% As a result, the model may assign high log-likelihoods to low-probability data, such as semantically implausible or chemically invalid structures.

We note that this issue is not unique to MDMs, but it does become significantly more severe in their case. Masking operations tend to absorb diverse input graphs into indistinguishable intermediate states, whereas substitution-based corruption in standard discrete diffusion preserves more structural variation and retains partial identity along the trajectory. \hj{Moreover, the effect is particularly pronounced in molecular graphs, which often contain structural symmetries and a limited set of element types compared to other discrete domains.}
% \textcolor{red}{Sungsoo: Add why does our framework solve state-clashing.}

\subsection{Algorithmic details}

\paragraph{\hj{Learnable element-wise embedding.}}
To ensure the forward diffusion trajectories among graph states to have as less probability of state-clashing, one should use information that distinguishes individual graph elements, which guides the noise scheduling network. One can consider incorporating graph positional encodings~\citep{lsp, grit} for conditioning. However, such encodings often fail to disambiguate elements when given symmetric structures such as those found in aromatic rings~\citep{sympe, orbitgnn}. Moreover, conditioning the noisy graph input in the noise schedule breaks the tractable closed-form marginal $q(\bm g_t\mid \bm g_0)$ since the transition kernel becomes dependent on the current corrupted state, which eliminates the efficiency.

Thus, we consider learnable element-wise embeddings over the graph elements that assigns distinct masking rate, and use it for an input to the noise scheduling network.  Specifically, we assign a learnable embedding matrix $\bm{H}\in\mathbb{R}^{D \times N}$ and consider its $i$-th column $\bm{h}^{i}$ as node-wise embedding of $i$-th node $x^i$, where $N>0$ is a number of nodes and $D$ is the embedding dimension. For an edge $\{i, j\}\in\mathcal{E}$, we set its embedding by $\bm{h}^{ij} = \bm{h}^{i} + \bm{h}^{j}$. In addition, we randomly permute columns of $\bm H$ during training to differentiate graph states that have the same numbers of nodes and edges.

\paragraph{\hj{Time-dependent noise schedule.}}
We parameterize the noise scheduling network for each element (\eg, node) using a power-law function, commonly used in \citet{md4, tabdiff}. Leveraging $i$-th node embedding $\bm{h}^{i}$ as an example, our noise schedule $\gamma_t^{i}$ is defined as:
\begin{equation}
\gamma_t^{i} = 1 - (1-\epsilon)\cdot t^{w^i},\quad w^i=\sigma_{\text{sf}}(\bm h^i),
\end{equation}
where $\sigma_{\text{sf}}$ denotes the softplus function. The same computation applies analogously to other nodes and edges. Consistent with \citet{md4, tabdiff}, we introduce a bounding constant $\epsilon$ for numerical stability and fix $\epsilon=0.0001$ in all experiments. Throughout this process, \system{} naturally introduces element-specific masking rates, mitigating the collapse between distinct molecules that would otherwise persist under unified noise scheduling.

\paragraph{Maintaining gradient flow in discrete sampling.}\label{subsec:optim} 
In discrete-space molecular diffusion frameworks~\citep{digress, graphdit, trainingfree_discrete}, the noisy graph at each timestep is obtained by sampling a single graph from a categorical distribution over nodes and edges (\cref{eq:cat_sampling}), as computing the full expectation over $\bm g_t\sim q(\cdot | \bm g)$ is intractable. However, such discretization introduces a discontinuity in the computational graph when parameterizing the forward process, impeding a gradient flow towards $q_\phi$. 

To circumvent this issue, we adopt the Straight-Through Gumbel-Softmax (STGS) trick~\citep{stgs}, which enables end-to-end training by providing a differentiable surrogate for discrete sampling. Let $\bm{z} \in \mathbb{R}^A$ denote the logits over $A$ possible node types, and $\eta > 0$ be the temperature parameter. We first compute a soft approximation of the categorical distribution $ \bm{p}_{\text{soft}} \in [0, 1]^{A} $ via the Gumbel-Softmax:
\begin{equation}
\bm{p}_{\text{soft}, k} = \frac{\exp((z_k + g_k)/\eta)}{\sum_{l=1}^{A} \exp((z_l + g_l)/\eta)},
\end{equation}
where $g_k = -\log(-\log(u_k)) $ is a gumbel noise with $u_k \sim \text{Unif}[0, 1]$ and $z_k$ is the $k$-th element of the logits $\bm{z}$. A discrete one-hot vector $\bm{p}_{\text{hard}} \in \{0, 1\}^{A}$ is then obtained by taking the index with the highest probability:
\begin{equation}
k^* = \arg\max_{k} \bm{p}_{\text{soft}, k}, \quad \bm{p}_{\text{hard}, k} = \begin{cases} 
1 & \text{if } k = k^* \\
0 & \text{otherwise}
\end{cases}
\end{equation}

To retain gradient flow, we use the straight-through estimator to combine the discrete and continuous components, \ie, set 
$\bm{p} = \bm{p}_{\text{hard}} - \text{sg}(\bm{p}_{\text{soft}}) + \bm{p}_{\text{soft}}$,
where $ \text{sg}(\cdot) $ denotes the stop-gradient operator. This ensures that the forward pass uses the discretized one-hot vector $ \bm{p} = \bm{p}_{\text{hard}}$, while the backward pass treats $\bm{p}$ as the continuous $\bm{p}_{\text{soft}}$, allowing gradients to propagate through $z$, \ie, $ \frac{\partial \bm{p}}{\partial z} = \frac{\partial \bm{p}_{\text{soft}}}{\partial z}$.

\begin{table}[t]
\centering
\caption{Unconditional generation of 10K molecules on QM9 and ZINC250K datasets. The best and second best performances are represented by \textbf{bold} and \underline{underline}.}
\label{tab:uncond_results}
\resizebox{\textwidth}{!}{
\begin{tabular}{l c c c c c c c c c c c c }
\toprule
& \multicolumn{6}{c}{QM9} & \multicolumn{6}{c}{ZINC250K} \\
\cmidrule(lr){2-7} \cmidrule(lr){8-13}
Method & Valid.$\uparrow$ & FCD$\downarrow$ & NSPDK$\downarrow$ & Scaf.$\uparrow$ & Uniq.$\uparrow$ & Novel.$\uparrow$ & Valid.$\uparrow$ & FCD$\downarrow$ & NSPDK$\downarrow$ & Scaf.$\uparrow$ &  Uniq.$\uparrow$ & Novel.$\uparrow$ \\
\midrule
\multicolumn{13}{l}{\emph{Flow-based}\vspace{0.02in}} \\
{\pz\pz}MoFlow & \phantom{0}91.36 & \pz4.47 & 0.017\pz & 0.145 & \underline{98.65} & \underline{94.72} & 
\phantom{0}63.11 & 20.93 & 0.046\pz & 0.013 & \phantom{0}\underline{99.99} & \textbf{100.00} \\
{\pz\pz}GraphAF & \phantom{0}74.43 & \pz5.63 & 0.021\pz & 0.305 & 88.64 & 86.59 & 
\phantom{0}68.47 & 16.02 & 0.044\pz & 0.067 & \phantom{0}98.64 & \phantom{0}\underline{99.99} \\
{\pz\pz}GraphDF & \phantom{0}93.88 & 10.93 & 0.064\pz & 0.098 & 98.58 & \textbf{98.54} & 
\phantom{0}90.61 & 33.55 & 0.177\pz & 0.000 & \phantom{0}99.63 & \textbf{100.00} \\
\arrayrulecolor{black!40}\midrule
\multicolumn{13}{l}{\emph{Continuous diffusion}\vspace{0.02in}} \\
{\pz\pz}EDP-GNN & \phantom{0}47.52 & \pz2.68 & 0.005\pz & 0.327 & \textbf{99.25} & 86.58 & 
\phantom{0}82.97 & 16.74 & 0.049\pz & 0.000 & \phantom{0}99.79 & \textbf{100.00} \\
{\pz\pz}GDSS & \phantom{0}95.72 & \pz2.90 & 0.003\pz & 0.698 & 98.46 & 86.27 & \phantom{0}97.01 & 14.66 & 0.019\pz & 0.047 & \phantom{0}99.64 & \textbf{100.00} \\
{\pz\pz}GruM & \phantom{0}\underline{99.69} & \pz{0.11} & \textbf{0.0002} & \underline{0.945} & 96.90 & 24.15 & 
\phantom{0}\underline{98.65} & \pz\underline{2.26} & \underline{0.0015} & \underline{0.530} & \phantom{0}{99.97} & \phantom{0}99.98 \\
\arrayrulecolor{black!40}\midrule
\multicolumn{13}{l}{\emph{Discrete diffusion}\vspace{0.02in}} \\
{\pz\pz}DiGress & \phantom{0}98.19 & \pz\underline{0.10} & \underline{0.0003} & 0.936 & 96.67 & 25.58 & 
\phantom{0}94.99 & \pz3.48 & 0.0021 & 0.416 & \phantom{0}{99.97} & \phantom{0}\underline{99.99} \\
\arrayrulecolor{black!40}\midrule
\multicolumn{13}{l}{\emph{Masked diffusion}\vspace{0.02in}} \\
{\pz\pz}GraphARM & \phantom{0}90.25 & \pz1.22 & 0.002\pz & N/A & 95.62 & 70.39 & 
\phantom{0}88.23 & 16.26 & 0.055\pz & N/A & \phantom{0}99.46 & \textbf{100.00} \\
{\pz\pz}MDM w/ cosine & \textbf{100.00} & \pz3.67 & 0.009\pz & 0.653 & 85.96 & 69.85 & 
\textbf{100.00} & 25.41 & 0.051\pz & 0.001 & \phantom{0}\underline{99.99} & \textbf{100.00} \\
{\pz\pz}MDM w/ polynomial & \textbf{100.00} & \pz3.70 & 0.010\pz & 0.890 & 86.57 & 67.18 & 
\textbf{100.00} & 26.43 & 0.053\pz & 0.001 & \phantom{0}99.93 & \textbf{100.00} \\
{\pz\pz}MDM w/ power-law & \textbf{100.00} & \pz3.62 & 0.007\pz & 0.628 & 91.30 & 76.65 & 
\textbf{100.00} & 26.09 & 0.068\pz & 0.001 & \textbf{100.00} & \textbf{100.00} \\
\rowcolor{lavender(web)}
\textbf{\pz\pz\system{} (Ours)}& \textbf{100.00} & \pz\textbf{0.09} & \textbf{0.0002}& \textbf{0.947} & 96.49 & 33.55 & 
\textbf{100.00} & \pz\textbf{1.51} & \textbf{0.0008} & \textbf{0.560} & \phantom{0}\underline{99.99} & \phantom{0}{99.95} \\
% \rowcolor{lavender(web)}
% \textbf{\pz\pz\system{} (Ours)}& \textbf{100.00} & \pz\textbf{0.10} & \textbf{0.0002}& \underline{0.941} & 96.89 & 37.40 & 
% \textbf{100.00} & \pz\textbf{1.82} & \textbf{0.0008} & \underline{0.517} & \textbf{100.00} & \textbf{100.00} \\
\arrayrulecolor{black}\bottomrule
\end{tabular}}
\vskip -10pt
\end{table}

\paragraph{\hj{Domain specialization and applicability.}}\label{subsec:applicability} 
In principle, \system{} is applicable to non-molecular data. However, we note that other discrete data such as text or protein sequences typically involve larger vocabularies and fewer structural symmetries. Consequently, the risk of collapsing distinct inputs into identical intermediate states is lower, and the relative benefits of \system{} may be less pronounced in such settings. Nevertheless, to show the generality of our approach, we include additional experiments on general graph with constrained number of nodes and edges in~\cref{4.4}.
\section{Experiments}\label{4_experiments}

\subsection{\hj{Experimental setup}}\label{4.1}
We evaluate \system{} on unconditional and property-conditioned molecular generation tasks. For unconditional generation, in line with prior work~\citep{grum,grapharm,gdss}, we use QM9~\citep{qm9}, ZINC250k~\citep{zinc}, and Guacamol~\citep{guacamol} datasets. For conditional generation, we adopt the Polymer dataset~\citep{polymer} introduced in~\citet{graphdit}, which conditions homopolymers on three gas permeability constraints and synthesizability scores. 
We compare against recent baselines with standard metrics for both tasks, following established setups~\citep{graphdit,gdss,grum}. See~\cref{C} for detailed description of each method and metrics. 
% Additionally, we compare our method with diverse noise schedules including both fixed (\cref{tab:uncond_results,tab:polymer_results} and learnable variants (\cref{tab:ablative_study}). 
Our implementation employs the diffusion transformer~\citep{dit} as the denoising network within a masked diffusion framework. For property-conditioned generation, we further apply classifier-free guidance~\citep{cfg} as implemented in~\citep{dit,graphdit}. Unless otherwise noted, all experiments use the same backbone across standard MDMs and \system{}.

\subsection{Main results}\label{4.2}
% In the following section, we present the results of \system{} upon benchmark datasets on unconditioned and property-conditioned generation tasks.

\paragraph{\hj{Unconditional Generation.}}
We present the results of \system{} on QM9 and ZINC250K datasets for unconditional generation. Remarkably, \system{} is the first approach to achieve 100\% chemical validity among conventional diffusion-based molecular generation methods, as shown in~\cref{tab:uncond_results}. On  the QM9 dataset, our method outperforms GraphARM~\citep{grapharm}, the autoregressive masked diffusion baseline, with up to 91\% reduction in FCD and NSPDK. Moreover, it substantially improves the distributional similarity by up to 98\% (in NSPDK) from standard MDMs.
% The low novelty observed in QM9, also reported in DiGress~\citep{digress} and GruM~\citep{grum}, is due to the limited generation space where molecules only consist of at most 9 atoms and 4 atom types (C, N, O, F). As noted in \citet{digress} (Appendix F), QM9 is an enumeration of small molecules satisfying given constraints. Thus, generating molecules outside this set does not reliably reflect the model's ability to capture the data distribution.} 

On the more challenging ZINC250K dataset, which includes larger molecules and richer atom types, \system{} achieves state-of-the-art results on 5 out of 6 metrics, surpassing GruM~\citep{grum}, the strongest baseline. It also consistently improves over masked diffusion baselines on key metrics including FCD, NSPDK, and scaffold similarity (Scaf.). In contrast, standard MDMs exhibit degenerate behavior, with FCD 91.4\% higher and a Scaf. 99.8\% lower than the best diffusion-based baselines, suggesting that unified schedulers yield valid but distributionally misaligned molecules.
% These results suggest that unified, fixed noise schedules yield valid but distributionally misaligned molecules, whereas our element-wise learnable schedule produces chemically valid molecules that better align with the target distribution.

\begin{table}[t]
    \centering\large
    \caption{Property-conditioned generation of 10K Polymers on three gas permeability properties and synthetic score. The numbers in parentheses in Valid. represent the validity without correction. The best and second best performances are represented by \textbf{bold} and \underline{underline}.}   
    \label{tab:polymer_results}
    \resizebox{\textwidth}{!}{
    \begin{tabular}{l c c c c c c c c c c}
    \toprule
     & \multicolumn{5}{c}{General Quality} & \multicolumn{5}{c}{Property Alignment} \\
     \cmidrule(lr){2-6} \cmidrule(lr){7-11}
     {\pz\pz}Method & Valid.$\uparrow$ & Cover.$\uparrow$ & Divers.$\uparrow$ & Frag.$\uparrow$ & FCD$\downarrow$ & Synth.$\downarrow$ & O$_2$ Perm.$\downarrow$ & N$_2$ Perm.$\downarrow$ & CO$_2$ Perm.$\downarrow$ & MAE$\downarrow$ \\
    \midrule
    \multicolumn{7}{l}{\emph{Molecule Optimization}\vspace{0.02in}} \\
    {\pz\pz}GraphGA            & 100.00 (N/A) & 11/11 & 88.28 & 0.927 & \pz9.19 & 1.3307 & 1.9840 & 2.2900 & 1.9489 & 1.888 \\
    {\pz\pz}MARS                & 100.00 (N/A) & 11/11 & 83.75 & 0.928 & \pz7.56 & 1.1658 & 1.5761 & 1.8327 & 1.6074 & 1.546 \\
    {\pz\pz}LSTM-HC             & \pz99.10 (N/A) & 10/11 & 89.18 & 0.794 & 18.16 & 1.4251 & 1.1003 & 1.2365 & 1.0772 & 1.210 \\
    {\pz\pz}JTVAE-BO            & 100.00 (N/A) & 10/11 & 73.66 & 0.729 & 23.59 & \textbf{1.0714} & 1.0781 & 1.2352 & 1.0978 & 1.121 \\
    \arrayrulecolor{black!40}\midrule
    \multicolumn{7}{l}{\emph{Continuous diffusion}\vspace{0.02in}} \\
    {\pz\pz}GDSS                & \pz92.05 (90.76) & \pz9/11  & 75.10 & 0.000 & 34.26 & 1.3701 & 1.0271 & 1.0820 & 1.0683 & 1.137 \\
    {\pz\pz}MOOD                & \pz98.66 (92.05) & 11/11 & 83.49 & 0.023 & 39.40 & 1.4019 & 1.4961 & 1.7603 & 1.4748 & 1.533 \\
    \arrayrulecolor{black!40}\midrule
    \multicolumn{7}{l}{\emph{Discrete diffusion}\vspace{0.02in}} \\
    {\pz\pz}DiGress v2          & \pz98.12 (30.57) & 11/11 & \textbf{91.05} & 0.278 & 21.73 & 2.7507 & 1.7130 & 2.0632 & 1.6648 & 2.048 \\
    {\pz\pz}GraphDiT     & \pz82.45 (84.37) & 11/11 & 87.12 & \underline{0.960} & \pz\underline{6.64} & 1.2973 & \underline{0.7440} & \underline{0.8857} & \underline{0.7550} & \underline{0.921} \\
    \arrayrulecolor{black!40}\midrule
    \multicolumn{7}{l}{\emph{Masked diffusion}\vspace{0.02in}} \\
    {\pz\pz}MDM w/ cosine & \pz15.95 (37.16) & 11/11 & \underline{89.91} & 0.307 & 26.45 & 2.1795 & 1.5035 & 1.7755 & 1.4974 & 1.739\\
    {\pz\pz}MDM w/ polynomial & \pz18.61 (60.32) & 11/11 & 88.44 & 0.237 & 29.32 & 2.0041 & 1.6805 & 1.9846 & 1.6468 & 1.829\\
    {\pz\pz}MDM w/ power-law & \pz17.31 (53.64) & 11/11 & 89.08 & 0.401 & 26.56 & 2.0145 & 1.4100 & 1.6536 & 1.4030 & 1.620 \\
    \rowcolor{lavender(web)}
    {\pz\pz}\textbf{\system{} (Ours)}& \pz99.10 (96.51) & 11/11 & 
    85.91 & \textbf{0.974} & \textbf{\pz5.93} & \underline{1.1398} & \textbf{0.6433} & \textbf{0.7596} & \textbf{0.6496} & \textbf{0.798} \\    
    \arrayrulecolor{black}\bottomrule
    \end{tabular}
    }
    \vskip -5pt
\end{table}

\paragraph{Property-conditioned Generation.} 
Next, we evaluate \system{} on conditional generation using the Polymer dataset~\citep{polymer}, with results summarized in~\cref{tab:polymer_results}. Overall, \system{} establishes a new state-of-the-art in property alignment, with a 13.4\% reduction in average MAE relative to GraphDiT~\citep{graphdit}. Apart from GraphDiT, no existing method can satisfy multiple property constraints simultaneously: LSTM-HC achieves strong synthesizability MAE but fails under gas permeability targets. DiGress v2~\citep{digress}, despite leveraging classifier guidance~\citep{classifier_guidance}, incurs substantially higher MAE across most conditions. Beyond alignment, \system{} also improves generative quality, surpassing FCD and fragment-based similarity (Frag.) over the previous best. Consistent with earlier work~\citep{cfg,tradeoff_nabla_gflownet}, we observe an inherent trade-off between property alignment and sample diversity. Importantly, our method addresses the state-clashing issue prevalent in MDMs: whereas element-agnostic schedule results in generating low-quality molecules, our learnable, element-wise noise schedule enhances validity by a factor of five and improves property alignment by an average of 50\%.

\begin{table}[t!]
%\vskip -10pt
\centering\small
\caption{Ablation study of \system{} with varying noise scheduling approaches. $\alpha$ without $\phi$ and $\alpha_\phi$ denote fixed and learnable schedules, respectively.}
\label{tab:ablative_study}
\resizebox{\textwidth}{!}{
\begin{tabular}{c l c c c c }
\toprule
Schedule type & Method & FCD$\downarrow$ & NSPDK$\downarrow$ & Scaf.$\uparrow$ & V.U.N.$\uparrow$\\
\midrule
\multirow{2}{*}{Fixed $\alpha$} 
& Power-law &  26.09 & 0.0683 & 0.001 & \textbf{100.00} \\
& DiffusionBERT~\citep{diffusionbert}
& \pz1.95 & {0.0009} & 0.491 & \textbf{100.00} \\
\arrayrulecolor{black}\midrule
\multirow{5.5}{*}{Learnable $\alpha_\phi$}
& GenMD4~\citep{md4} & \pz3.19 & 0.0017 & 0.429 & \textbf{100.00} \\
& TabDiff~\citep{tabdiff} & \pz2.15 & {0.0009} & 0.486 & \phantom{0}{99.99} \\
\arrayrulecolor{black!30}\cmidrule(lr){2-6}
& \system{} \textbf{(Ours; Node)}& \pz1.63 & 0.0009 & 0.536 & \phantom{0}99.99 \\
& \system{} \textbf{(Ours; Edge)} & \pz1.73 & 0.0009 & 0.525 & \phantom{0}99.99 \\
& \cellcolor{lavender(web)}\system{} \textbf{(Ours; Node + Edge)} & \cellcolor{lavender(web)}\pz\textbf{1.51} & \cellcolor{lavender(web)}\textbf{0.0008} & \cellcolor{lavender(web)}\textbf{0.560} & \cellcolor{lavender(web)}\phantom{0}{99.94} \\
\arrayrulecolor{black}\bottomrule
\end{tabular}
}
\vskip -10pt
\end{table}

\begin{figure}[t!]
    \centering
    \includegraphics[width=\linewidth]{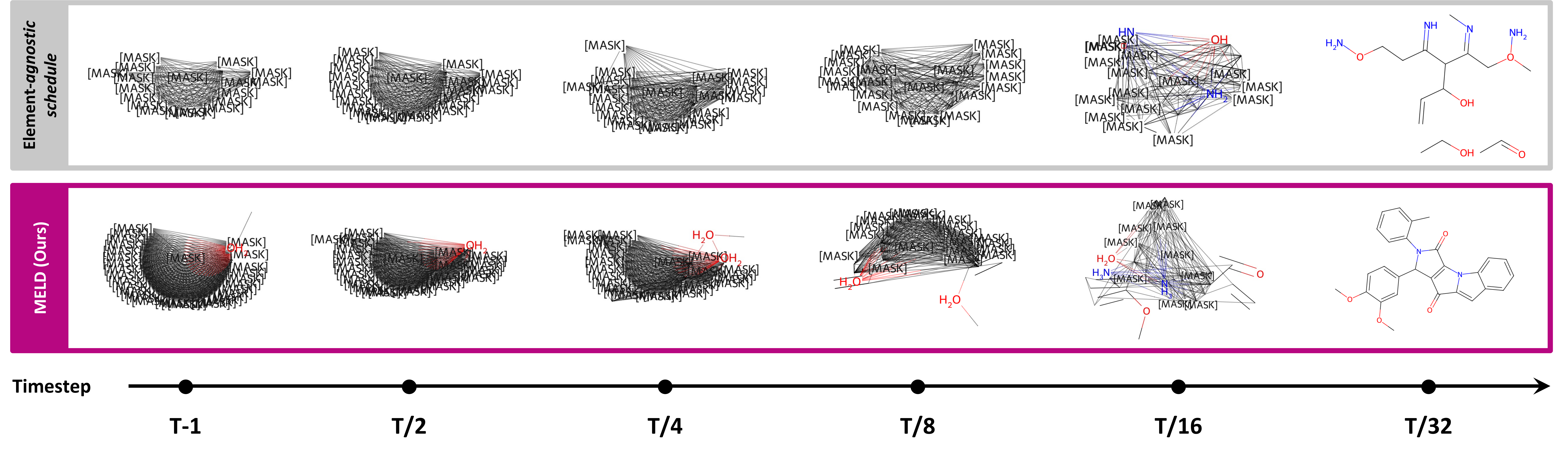}
    % \vspace{-0.2in}
    \caption{
    Comparison between fixed power-law scheduling and \system{} during reconstruction. With the learnable noise schedule, \system{} achieves faster recovery than standard MDMs.
    }
    \vskip -10pt
    \label{fig:analysis_reverse_process}
\end{figure}
% \vspace{-5mm}
\vspace{-1mm}
\subsection{\hj{Ablation study}}\label{4.3}
We evaluate several learnable scheduling strategies on ZINC250K~\citep{zinc}, as summarized in~\cref{tab:ablative_study}. The first row reports a standard MDM with a unified power-law function, while the second and third rows correspond to element-wise learnable scheduling applied only to nodes or edges. Rows four through six present advanced scheduling strategies from prior work. DiffusionBERT~\citep{diffusionbert} employs a fixed spindle noise schedule decided by class-wise entropy; GenMD4~\citep{md4} is another class-wise scheduling variant where each atom and bond type has its own learned corruption rate; and TabDiff~\citep{tabdiff} introduces a single corruption rate shared across elements within the same column, analogous to node and edge-level schedules, \eg, all nodes sharing the same schedule. The final row corresponds to the full element-wise schedule of \system{}.

As depicted in the table, all alternative noise schedules fall short of optimal gains in key metrics such as FCD and Scaf., an effect we attribute to their limited ability of reducing state-clashing. For instance, employing GenMD4 noise scheduling can remain limited in resolving the state-clashing: delaying the corruption of all carbon atoms relative to nitrogen in o-Phenylenediamine (\cref{fig:framework}) may still result in symmetric benzene ring. In contrast, our full per-element corruption (\system{}) delivers further reductions in distributional similarity metrics, demonstrating its fine-grained control. 
% demonstrating that fine-grained control over the diffusion trajectory better distinguishes semantically similar elements. This further highlights the effectiveness of proposed noise scheduling for target-aligned molecular generation.

\vspace{-1mm}
\subsection{Qualitative analysis}\label{4.4}

\paragraph{\hj{Reverse process of \system{}.}} 
\cref{fig:analysis_reverse_process} compares \system{} with standard MDM. Corrupted nodes and edges are shown as $[\texttt{mask}]$ and dashed lines, respectively. Under unified noise scheduling, unmasking proceeds relatively slowly: at time $t=T/16$ nitrogen and oxygen functional groups only begin to emerge, ultimately leading to a poorly-designed molecule. In contrast, \system{} reconstructs fragments earlier relative to element-agnostic schedule, where H$_2$O and its surrounding bonds already reveal at step $t=T-1$. Similar phenomena can also be found with more examples in~\cref{D:more_reverse_process}. 

\vspace{-1mm}
\paragraph{\hj{Scalability to large molecules.}}
\begin{wraptable}[9]{r}{0.56\textwidth}
\centering\small
\vskip -13pt
\caption{Performance comparison on large-scale Guacamol dataset. The metrics are transformed such that higher values indicate better performance.}
\label{tab:guacamol_results}
\resizebox{.55\textwidth}{!}{
\begin{tabular}{l c c c c c }
\toprule
Method & Valid.$\uparrow$ & Unique.$\uparrow$ & Novel.$\uparrow$ & KL div$\uparrow$ & FCD$\uparrow$  \\
\midrule
ConGress~\citep{digress} & \phantom{0}\phantom{0}0.1 & \textbf{100.0} & \textbf{100.0} & 36.1 & \phantom{0}0.0  \\
DiGress~\citep{digress} & \phantom{0}85.2 & \textbf{100.0} & \phantom{0}99.9 & 92.9 & 68.0 \\
DisCo~\citep{disco} & \phantom{0}86.6 & 100.0 & \phantom{0}99.9 & 92.6 & 59.7  \\
\rowcolor{lavender(web)}\system{} & \textbf{100.0} & \textbf{100.0} & \textbf{100.0} & \textbf{93.4} & \textbf{68.8} \\
\arrayrulecolor{black}\bottomrule
\end{tabular}}
% \vskip -10pt
\end{wraptable}

We further evaluate \system{} on the large-scale Guacamol dataset~\citep{guacamol} following the standard protocol used in prior work~\citep{digress}. As demonstrated in~\cref{tab:guacamol_results}, \system{} surpasses all diffusion-based baselines~\citep{digress,disco} and is the first to achieve 100\% validity among molecular diffusion models. Notably, this performance is obtained with 70\% reduced training epochs (300 epochs) than DiGress (1000 epochs), emphasizing both efficiency and empirical gains.

\vspace{-1mm}
\begin{wraptable}[9]{r}{0.56\textwidth}
\vskip -13pt
\centering
\caption{Number of unique graph states across varying timesteps in ZINC250K, averaged over 3 seeds.}
\label{tab:unique_graphs_zinc}
\resizebox{.55\textwidth}{!}{%
\small
\begin{tabular}{l c c c c c }
\toprule
Method & T-100 & T-75 & T-50 & T-25 & T-1 \\
\midrule
MDM w/ cosine & \textbf{131.0} & 122.3 & \pz63.0 & \pz14.7 & \pz1.7 \\
MDM w/ polynomial & \textbf{131.0} & \textbf{131.0} & \textbf{131.0} & 103.0 & 13.3 \\
MDM w/ power-law & \textbf{131.0} & \textbf{131.0} & \textbf{131.0} & 126.0 & \pz8.7 \\
\rowcolor{lavender(web)}\system{} & \textbf{131.0} & \textbf{131.0} & \textbf{131.0} & \textbf{131.0} & \textbf{17.3} \\
\arrayrulecolor{black}\bottomrule
\end{tabular}
}
% \vskip -5pt
\end{wraptable}
\paragraph{\hj{Quantifying state-clashing problem.}}
Here, we assess state-clashing phenomenon by measuring the number of distinct intermediate graph states at each timestep, as shown in~\cref{tab:unique_graphs_zinc}. Specifically, we sample molecules with a fixed graph size and employ a graph isomorphism-based method~\citep{isomorphism} to count unique graphs. A higher count of indicates fewer state-clashing. Due to the extreme cost of isomorphism algorithm, we sample 131 molecules with 12 nodes from the ZINC250K dataset for the evaluation. The results show that \system{} preserves greater structural diversity at later timesteps compared to any standard MDMs.

It is important to note that \system{} is not intended to eliminate state-clashing \textit{entirely}, but to reduce the chance of its occurrence, particularly in the early and intermediate timesteps. Inevitably, some clashes remain, \eg, all graphs converge to a fully masked state, but these unavoidable cases only affects a small portion of decisions near the prior distribution and therefore does not compromise its overall effectiveness.

% Furthermore, we clarify that although theoretically all modalities share the problem of state-clashing, it is most problematic in the molecular graph regime. This is due to the limited vocabulary size (\eg, 4 atom types in QM9), and frequent symmetries (\eg, benzene ring). For these reasons, \system{} is most beneficial within molecular graph generation.

\vspace{-1mm}
\paragraph{\hj{Generalizability on synthetic graph.}}
% \begin{table}[t!]
\begin{wraptable}[8]{r}{0.56\textwidth}
\centering\small
\vskip -13pt
\caption{Performance comparison on synthetic graph domain (SBM).}
\label{tab:sbm_results}
\resizebox{.55\textwidth}{!}{
\begin{tabular}{l c c c c c }
\toprule
Method & Degree$\downarrow$ & Cluster$\downarrow$ & Orbit$\downarrow$ & Spectral$\downarrow$ & V.U.N.$\uparrow$ \\
\midrule
DiGress & 0.0013 & 0.0498 & 0.0434 & 0.0400 & 74.00 \\
GruM & 0.0007 & \textbf{0.0492} & 0.0448 & 0.0050 & 85.00 \\
\rowcolor{lavender(web)}\system{} & \textbf{0.0005} & 0.0506 & \textbf{0.0381} & \textbf{0.0047} & \textbf{97.50} \\
\arrayrulecolor{black}\bottomrule
\end{tabular}
}
% \vskip -5pt
\end{wraptable}
% \end{table}

% \input{tables/unique_graphs_sbm}
% While the state-clashing can also arise in other discrete graph domains such as citation graphs~\citep{bernecker2024random} or social networks~\citep{ji2024llm}, they typically involve a more diverse set of node and edge types, which reduces the likelihood that distinct graphs collapse into identical intermediate states. However, in synthetic graphs such as SBM~\citep{sbm}, which contain only a single node type and binary edge types (denoting edge existence), state-clashing is susceptible to occur as depicted in~\cref{tab:unique_graphs_sbm} in~\cref{D:sbm}.

To assess generalizability of \system{} on other discrete graph domains, we benchmark \system{} against two strong molecular diffusion models, DiGress and GruM, on SBM~\citep{sbm}, a synthetic graph benchmark. Following the standard evaluation protocol~\citep{digress,grum}, we compute the maximum mean discrepancy (MMD) across four key graph statistics. As reported in~\cref{tab:sbm_results}, \system{} outperforms the baselines on most metrics, with particularly notable gains in V.U.N. and Orbit.
\section{Conclusion}\label{6_conclusion}

In this work, we investigated masked diffusion models (MDMs) for molecular graph generation and identified a central limitation, which we term \emph{state-clashing}. To address this, we introduced \system{}, a masked diffusion model that learns element-wise forward trajectories through a parameterized noise scheduling. Extensive experiments show that \system{} consistently outperforms standard MDMs and prior diffusion-based methods in both unconditional and property-conditioned molecular generation.

\subsubsection*{Acknowledgments}
This work was partly supported by Institute for Information \& communications Technology Planning \& Evaluation(IITP) grant funded by the Korea government(MSIT) (RS-2019-II190075, Artificial Intelligence Graduate School Support Program(KAIST)), National Research Foundation of Korea(NRF) grant funded by the Ministry of Science and ICT(MSIT) (No. RS-2022-NR072184), GRDC(Global Research Development Center) Cooperative Hub Program through the National Research Foundation of Korea(NRF) grant funded by the Ministry of Science and ICT(MSIT) (No. RS-2024-00436165), the Institute of Information \& Communications Technology Planning \& Evaluation(IITP) grant funded by the Korea government(MSIT) (RS-2025-02304967, AI Star Fellowship(KAIST)), and Polymerize.

\bibliography{iclr2026_conference}
\bibliographystyle{iclr2026_conference}

\newpage
\appendix

\appendix
\section*{Supplementary Materials}

\section{More related work}\label{A}
\paragraph{Molecule optimization.}
Optimization-based methods generate molecules by iteratively refining candidates assembled from a predefined vocabulary of fragments, aiming to align with desired property constraints. These approaches typically employ techniques such as genetic algorithms~\citep{graphga}, Bayesian optimization~\citep{bayesian_opt, jtvae, sample_gflownet}, and goal-directed generation~\citep{goal_directed, graph_goal}. Representative examples include~\citep{rationale, jtvae, mars, differentiable}, which utilize predefined subgraph motifs or scaffolds to ensure chemical validity during the generation process. These methods rely on diverse strategies including Markov sampling to sparse Gaussian processes and optimize molecules based on property-specific scoring functions. Goal-directed generation~\citep{goal_directed, graph_goal}, in particular, often adopts reinforcement learning, where a generation policy is updated to maximize a property-driven reward function. Despite their strengths, existing optimization-based approaches remain limited in conditional generation settings. Specifically, they require a full re-optimization for each new property configuration when tasked with generating molecules that precisely match target properties, rather than simply increasing or decreasing property values. This results in a high training complexity and limits their scalability~\citep{xmol, opt_survey}.

\paragraph{Learnable noise scheduling.}
Several works have explored learnable corruption process to optimize the forward trajectories in images and text. In continuous-space diffusion models, \citet{vdm} introduces a learnable scalar noise schedule as a function of time, enabling variance reduction in evidence lower bound (ELBO) estimation. Extending this, \citet{mulan} proposes a multivariate, data-dependent noise schedule, showing that a non-scalar, adaptive diffusion process can further tighten the ELBO by aligning the forward process more closely with the true posterior. In discrete masked diffusion, \citet{md4} generalizes the corruption process to allow class-dependent masking rates across tokens, prioritizing semantically important tokens during generation. \citet{tabdiff} adopts feature-wise noise schedule for tabular data, where a single corruption rate is shared across elements within the same column. Additionally, Schrödinger bridges-based approaches~\citep{dsb1, dsb2, dsb3} formulate generative modeling as learning an expressive, path-wise forward process by solving entropy-regularized optimal transport problems over path spaces.

It is noteworthy that the design philosophy of \system{} is built upon the state-clashing, a critical issue that has not been addressed in these work nor in the molecular diffusion literature~\citep{digress, gdss, grum, graphdit, mood}. While employing the learnable forward process, our work departs from existing methods by introducing graph element-wise parameterization of the forward diffusion, specifically to avoid trajectory collisions between semantically distinct molecules. Moreover, we explicitly target and resolves the intermediate state degeneracy unique to discrete molecular graphs, while Schr\"odinger bridge-based approaches neither address structural collapse in discrete settings nor differentiate forward paths across individual graph elements.

\section{Limitations}\label{B}
While our element-wise noise scheduling significantly mitigates the state-clashing issue, it may not fully address the inherent multimodality when a large portion of molecules are masked at later diffusion steps. 
This is especially pronounced at later diffusion steps, where a majority of the graph elements are masked, making it challenging to distinguish them. Nevertheless, these unavoidable cases only affects a small portion of corruption near the prior distribution and therefore does not compromise the overall efficacy of our method.

\section{Experimental setup}\label{C}
\paragraph{Implementation details.}
We follow the evaluation protocols and dataset splits adopted in prior works: for unconditional generation, we adopt the setup from \cite{grum}, and for property-conditioned tasks, we follow the procedure outlined in \cite{graphdit}. We provide the detailed statistics of each dataset in~\cref{tab:statistics}. During training for unconditional generation, we apply an exponential moving average (EMA) to the model parameters, consistent with the training framework in \cite{grum}. For conditional generation, we utilize the implementation strategies proposed in \cite{dit, graphdit}, including condition vector encoders and adaptive layer normalization (AdaLN). All models are implemented in PyTorch~\cite{NEURIPS2019_pytorch} with PyTorch Geometric~\cite{2019torch_geometric}. Experiments were conducted on machines equipped with NVIDIA RTX 4090 and A5000 GPUs (24 GB) and AMD EPYC 7543 32-Core CPUs (64 cores total). Across all experiments, we use a transformer-based denoising model~\citep{dit} with 6 layers, a hidden dimension of 1152, and 16 attention heads. The noise scheduling network is parameterized as a two-layered MLP with SiLU activation with hidden dimension set as 64. We train all models using the AdamW optimizer with batch sizes between [512, 1024], no weight decay, and a fixed random seed of 0.
\begin{table}[ht]
\centering\small
% \vskip -10pt
\caption{Dataset statistics.}
\label{tab:statistics}
%\resizebox{.8\textwidth}{!}{
\begin{tabular}{l c c c c }
\toprule
Dataset & \#(Graphs) & \#(Nodes) & \#(Node types) & \#(Edge types) \\
\midrule
QM9 & 133,985 & $|\gV|\le$ 9 & 4 & 3 \\
ZINC250K & 249,555 & $|\gV|\le$ 38 & 9 & 3 \\
Polymers & 553 & $|\gV|\le 50$ & 11 & 3 \\
\arrayrulecolor{black}\bottomrule
\end{tabular}
%}
\vskip -10pt
\end{table}

\paragraph{Baselines.}
We consider various recent baselines for conditional and unconditional generation; following experimental setups of prior works~\citep{graphdit, gdss, grum}. 
\begin{itemize}[topsep=0pt,itemsep=1mm, parsep=0pt, leftmargin=5mm]
\item \textbf{Unconditional Generation}: First, we consider three flow-based models as baselines: MoFlow~\citep{zang2020moflow}, GraphAF~\citep{graphaf} and GraphDF~\citep{graphdf}, three continuous diffusion models: EDP-GNN~\citep{edp-gnn}, GDSS~\citep{gdss}, and GruM~\citep{grum}, and one discrete-diffusion model: Digress~\citep{digress}. Additionally, we compare ~\system{} against GraphARM~\citep{grapharm}, a method that employs mask tokens as absorbing states but generates tokens (\ie nodes) autoregressively. 
\item \textbf{Conditional Generation}: We consider four optimization-based frameworks as baselines: GraphGA~\citep{graphga}, MARS~\citep{mars}, LSTM-HC~\citep{neil2018exploring}, and JTVAE-BO~\citep{jtvae}, two continuous diffusion models: GDSS~\citep{gdss} and MOOD~\citep{mood}, and two discrete diffusion models: DiGress v2~\citep{digress} integrated with classifier guidance and GraphDiT~\citep{graphdit}. 
\end{itemize}

\paragraph{Metrics.}
Following the evaluation protocol in previous work~\citep{graphdit, gdss, grum}, we evaluate the performance of our framework using the following metrics:
\begin{itemize}[topsep=0pt,itemsep=1mm, parsep=0pt, leftmargin=5mm]
\item \textbf{Unconditional Generation}: We use 10,000 generated samples for evaluation using the following six metrics: (1) \textit{Valid.}, the proportion of chemically valid molecules; (2) \textit{Frechet ChemNet Distance} (FCD;~\citealt{fcd}), a distributional similarity score of ChemNet embeddings between generated and reference molecules; (3) \textit{NSPDK}~\citep{nspdk}, a graph kernel metric that quantifies topological similarity to the reference set; (4) \textit{Scaf.}, a scaffold-level similarity score; (5) \textit{Uniqueness}, the proportion of valid molecules that are structurally distinct within the generated set; and (6) \textit{Novelty}, the fraction of valid molecules not in the training data.

\item \textbf{Conditional Generation}: We generate 10,000 samples and assess their overall quality using the following criteria: (1) \textit{Valid.}, (2) \textit{Cover.}, the heavy atom type coverage; (3) \textit{Divers.}, the diversity among the generated molecules; (4) \textit{Frag.}, a fragment-based similarity metric; and (5) FCD. We also report \textit{Property Alignment}, measured as the mean absolute error (MAE) between target properties and the corresponding oracle-evaluated scores of generated molecules. 
\end{itemize}
To compute property alignment, we follow the setup of prior works~\citep{graphdit, gao2022sample}, employing a random forest model trained on molecular fingerprints as an oracle function. 

We collect the values of each baseline reported from prior works~\citep{gdss, graphdit} for both settings. To evaluate the efficacy of our method in remedying state-clashing, we perform additional ablative studies against fixed-scheduling mechanisms often adopted in masked diffusion models; namely cosine (MDM w/ cosine), polynomial (MDM w/ polynomial), and power-law (MDM w/ power-law) scheduling functions.

\section{Further experiments and analysis}\label{D}

\subsection{\hj{Analysis of element-wise learned embedding}}\label{D:symmetry}

\begin{wraptable}[7]{r}{0.56\textwidth}
\centering\small
\vskip -13pt
\caption{Average cosine similarity between pairs of $\bm h^i$ and $\bm h^{ij}$ in a benzene ring.}
\vskip -5pt
\label{tab:symmetry_results}
\resizebox{.55\textwidth}{!}{
\begin{tabular}{l c c }
\toprule
Cosine similarity & Learned $\bm H$ (\system{}) & Random walk embedding \\
\midrule
Nodes ($\bm h^i$) & 0.103 & 1 \\
Edges ($\bm h^{ij}$) & 0.237 & 1 \\
All & 0.192 & 1 \\
\arrayrulecolor{black}\bottomrule
\end{tabular}
}
% \vskip -5pt
\end{wraptable}

Our design philosophy of learnable embedding in \system{} is focused on reducing the chance of state-clashing problem by making each graph elements distinct and unique. As a result, our method can distinguish graph elements even within the symmetric motifs, which is often difficult to be discriminated using existing graph positional encodings~\citep{lsp, grit}.

To empirically verify this, we analyze the learned embedding matrix $\bm H$ on a benzene ring and compare it to random walk positional embeddings~\citep{lsp}. In~\cref{tab:symmetry_results}, we evaluate the average pairwise cosine similarity for (1) node embeddings $\bm h^i$, (2) edge embeddings $\bm h^{ij}$ (connected to benzene ring), and (3) all element embeddings (nodes and edges). Our learned embeddings exhibit significantly low pair-wise similarity, suggesting that the learned embedding successfully distinguishes elements even within the symmetric structure.

\subsection{Per-element scheduling of \system{}}\label{D:cv}

\begin{wrapfigure}[10]{r}{0.35\textwidth}
\vskip -13pt
\centering
    \includegraphics[width=.95\linewidth]{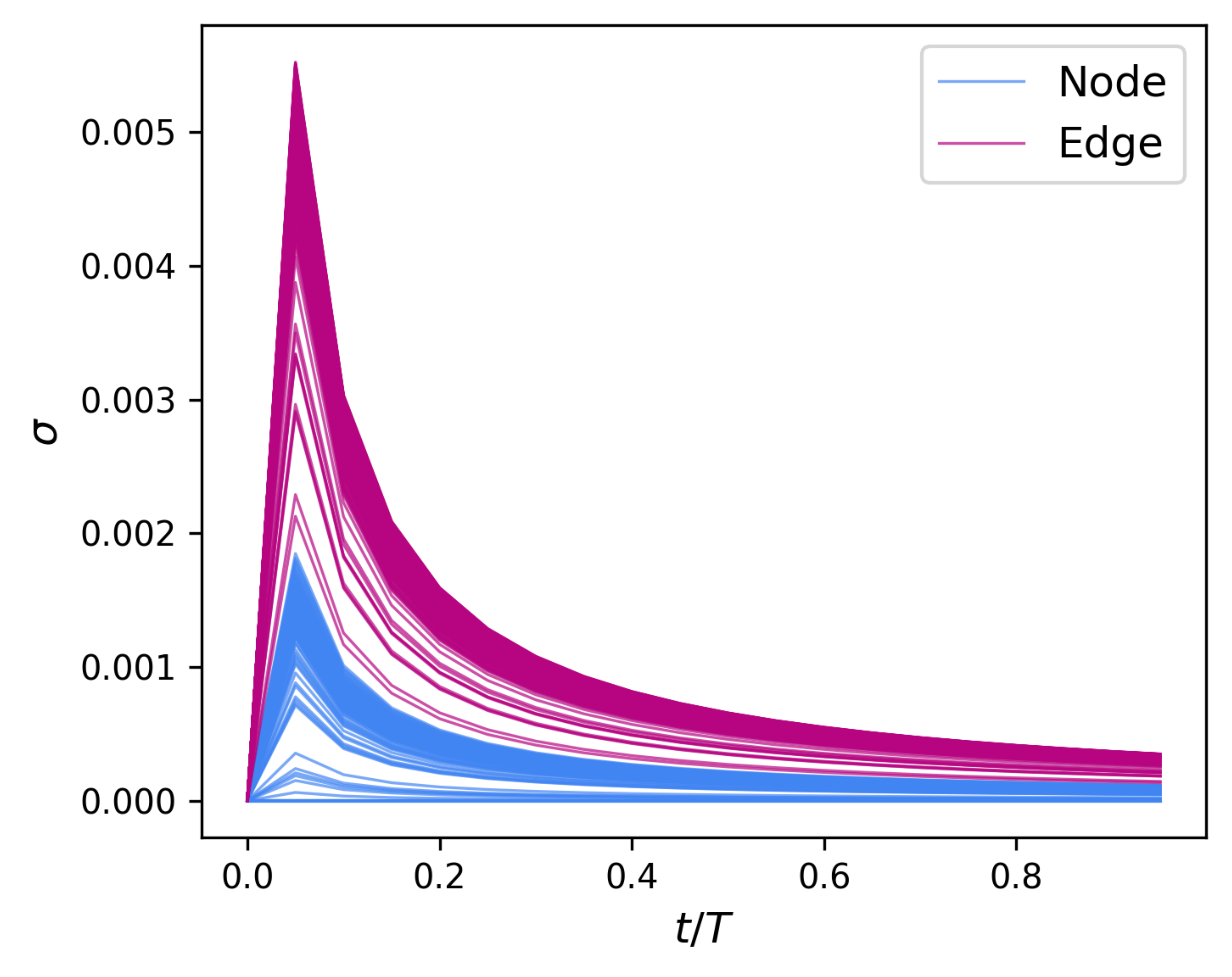}
        \vskip -10pt
        \caption{Variation of normalized masking probability $\sigma$.}
        \label{fig:analysis_cv}
        % \vskip 20pt
\end{wrapfigure}

In~\cref{fig:analysis_cv}, we visualize the variation in per-step learned noise schedules across nodes and edges during the forward diffusion process. Specifically, we take 200 samples and plot the variation of the normalized masking probability $\sigma$, defined as the standard deviation of $\smash{\frac{\alpha_{t-1,\phi} - \alpha_{t,\phi}}{1 - \alpha_{t,\phi}}}$. 

We observe consistently higher variance for edge schedules across all timesteps, suggesting that the model prioritizes differentiating edges more aggressively than nodes during training. In addition, state-clashing problem is inherently intensified in the later steps of the forward process for both nodes and edges, as expected. 

% \vspace{10mm}
\begin{figure}[t!]
\centering
\includegraphics[width=\linewidth]{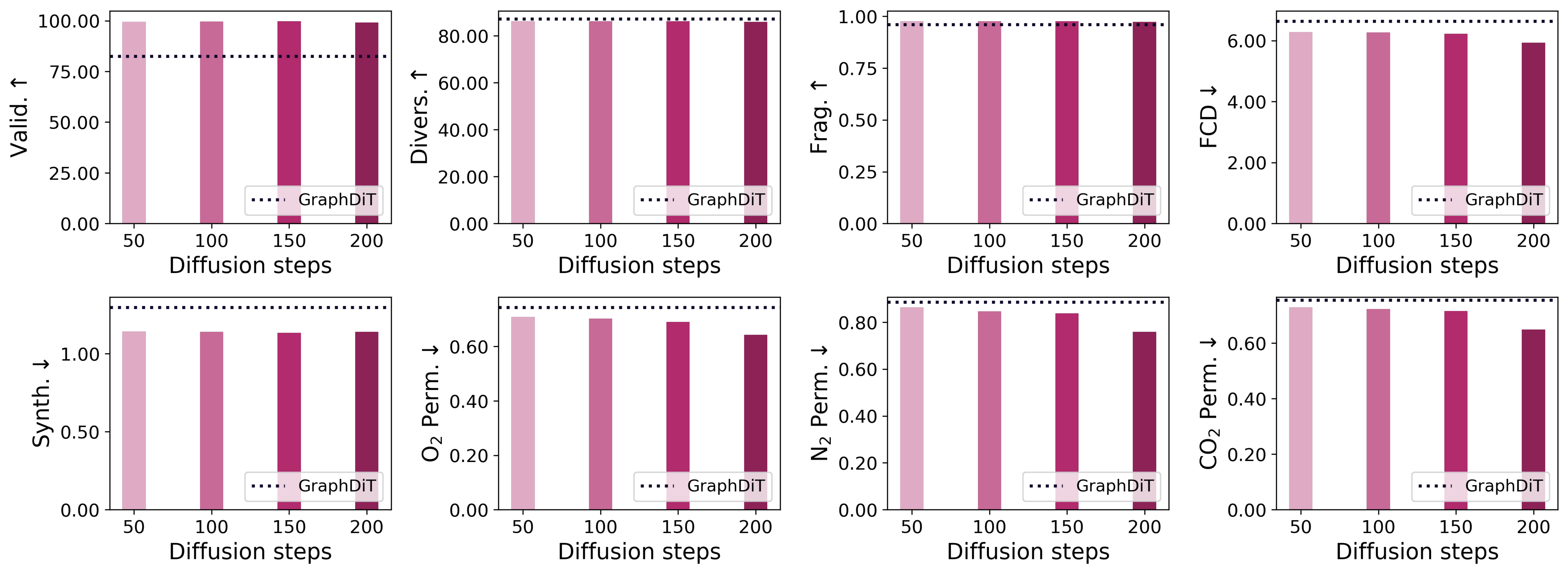}
% \vspace{-0.2in}
\caption{
Performance of \system{} under varying diffusion steps. The dotted line indicates the performance of the strongest baseline, GraphDiT, evaluated at a fixed diffusion step of 500.
}
% \vspace{-0.15in}
\label{fig:appendix_diffusion_step}
\vskip -7pt
\end{figure}

\subsection{Robustness across different diffusion steps}\label{D:step_robustness}
We evaluate the performance of \system{} on the Polymer dataset under varying diffusion steps, setting the total timestep $T\in\{50,100,150,200\}$. Note that during this experiment, we fix the \system{}-incorporated MDM to be trained upon a fixed diffusion step of $200$, and only vary the number of steps taken during inference. We compare the performance of \system{} against that of the strongest baseline, GraphDiT~\citep{graphdit}, which is originally evaluated at diffusion step size of 500. As depicted in~\cref{fig:appendix_diffusion_step}, \system{} overall exhibits robust performance across a range of metrics. 
% Interestingly, for validity, diversity and fragment similarity and synthesizability, we see consistent performance across diffusion steps. However, for FCD, O$_2$, N$_2$, and CO$_2$ permeability, we see monotonic gains in performance as we increase the diffusion step size.  

% \paragraph{Robustness across hyperparameters introduced by \system{}.}
% In addition to the number of diffusion steps, we provide sensitivity analysis of noise scheduling bounding values $\gamma_\min, \gamma_\max$ and STGS temperature $\eta$ in~\cref{}. Note that $\gamma_\min$ and $\gamma_\max$ are used to ensure \textit{numerical stability} after the sigmoid operation. Unless the values become extreme, \eg, where $\gamma_\min=-1$ and $\gamma_\max=1$ cause $\alpha_{0,\phi}^i=0.73$ and $\alpha_{1,\phi}^i=0.27$, we find the performance is maintained. Similarly, \system{} shows minimal performance variations across diverse $\eta$.

% \subsection{Results on large-scale dataset}\label{D:guacamole}
% \input{tables/table_guacamol}
% We present additional experimental results on Guacamol~\citep{guacamol}, a large molecular dataset used in DiGress~\citep{digress}, in~\cref{tab:guacamol_results}. We adopt the same evaluation metrics as in DiGress~\citep{digress} to ensure fairness. \hj{TODO: add description after finishing the exp.}

\subsection{\hj{State-clashing on synthetic graph}}\label{D:sbm}
\begin{wraptable}{r}{0.56\textwidth}
\centering\small
\vskip -13pt
\caption{Number of unique graph states across varying timesteps in synthetic graph domain (SBM), averaged over 3 seeds.}
% \vskip -5pt
\label{tab:unique_graphs_sbm}
\resizebox{.55\textwidth}{!}{
\begin{tabular}{l c c c c c c}
\toprule
Method & T-100 & T-75 & T-50 & T-25 & T-1 \\
\midrule
MDM w/ cosine & 36.7 & 23.3 & 11.7 & \pz4.3 & 1.3 \\
MDM w/ polynomial & \textbf{72.0} & \textbf{72.0} & 58.0 & 18.7 & \textbf{7.3} \\
MDM w/ power-law & \textbf{72.0} & 71.3 & 66.0 & 38.6 & 4.0 \\
\rowcolor{lavender(web)}\system{} & \textbf{72.0} & \textbf{72.0} & \textbf{72.0} & \textbf{65.7} & 6.3 \\
\arrayrulecolor{black}\bottomrule
\end{tabular}
}
\vskip -5pt
\end{wraptable}
While the state-clashing can also arise in other discrete graph domains such as citation graphs~\citep{bernecker2024random} or social networks~\citep{ji2024llm}, they typically involve a more diverse set of node and edge types, which reduces the likelihood that distinct graphs collapse into identical intermediate states. However, in synthetic graphs such as SBM~\citep{sbm}, which contain only a single node type and binary edge types (denoting edge existence), state-clashing is susceptible to occur as depicted in~\cref{tab:unique_graphs_sbm}.

Using SBM synthetic graph~\citep{sbm} as a representative, we conducted the same quantitative analysis as done in~\cref{tab:unique_graphs_zinc}. Due to the high computational cost of applying a full graph isomorphism check on the original graphs, we adopted a practical approximation: for each of 72 test and validation graphs, we randomly sampled 10 nodes and performed the state-clashing analysis. This procedure was repeated over 3 random seeds to ensure consistency. The results show that \system{} mostly outperforms standard MDMs in terms of distinguishability.

\subsection{\hj{Computational cost analysis}}\label{D:cost}
\begin{table}[t!]
\centering\small
\caption{Computational cost analysis with varying molecular sizes. 
All experiments were conducted on an NVIDIA GeForce RTX 4090 GPU and 
an AMD EPYC 7K62 48-Core Processor. Runtime values are averaged over 5 random seeds.}
\label{tab:cost_results}
\begin{tabular}{c c c c c c}
\toprule
\#Atoms & Method & \#Params (M) & FLOPS (GMac) & Peak memory (MB) & Runtime (sec) \\
\midrule
\multirow{2}{*}{10} & \pz MDM  & 156.23 & \phantom{0}2.06 & \phantom{0}622.76 & 0.056 $\pm$ 0.005 \\
                    & \system{} & 156.24 & \phantom{0}2.06 & \phantom{0}623.19 & 0.056 $\pm$ 0.006 \\
\multirow{2}{*}{50} & \pz MDM  & 157.22 & \phantom{0}9.88 & \phantom{0}680.40 & 0.077 $\pm$ 0.007 \\
                    & \system{} & 157.23 & \phantom{0}9.93 & \phantom{0}692.31 & 0.087 $\pm$ 0.008 \\
\multirow{2}{*}{100}& \pz MDM  & 158.46 & 19.74 & \phantom{0}755.12 & 0.093 $\pm$ 0.008 \\
                    & \system{} & 158.47 & 19.94 & \phantom{0}788.71 & 0.098 $\pm$ 0.008 \\
\multirow{2}{*}{200}& \pz MDM  & 160.93 & 39.77 & \phantom{0}929.09 & 0.132 $\pm$ 0.007 \\
                    & \system{} & 160.94 & 40.59 & 1067.64 & 0.165 $\pm$ 0.008 \\
\bottomrule
\end{tabular}
\vskip -10pt
\end{table}
In practice, the computational and memory overhead introduced by \system{} is negligible since it only adds learnable embedding matrix $\bm H$ to the existing transformer-based architectures. To validate this, we report (1) total number of parameters, (2) FLOPs, (3) peak memory usage, and (4) single-step runtime across various molecular sizes ($|\mathcal V|\in[10,50,100,200]$) for \system{} and standard MDM in~\cref{tab:cost_results}. Our results demonstrate that \system{} introduces only about 0.01M additional parameters with negligible computational overhead, regardless of the input size.

\subsection{More examples of reverse process}\label{D:more_reverse_process}
\begin{figure}[t!]
    \centering
    \includegraphics[width=\linewidth]{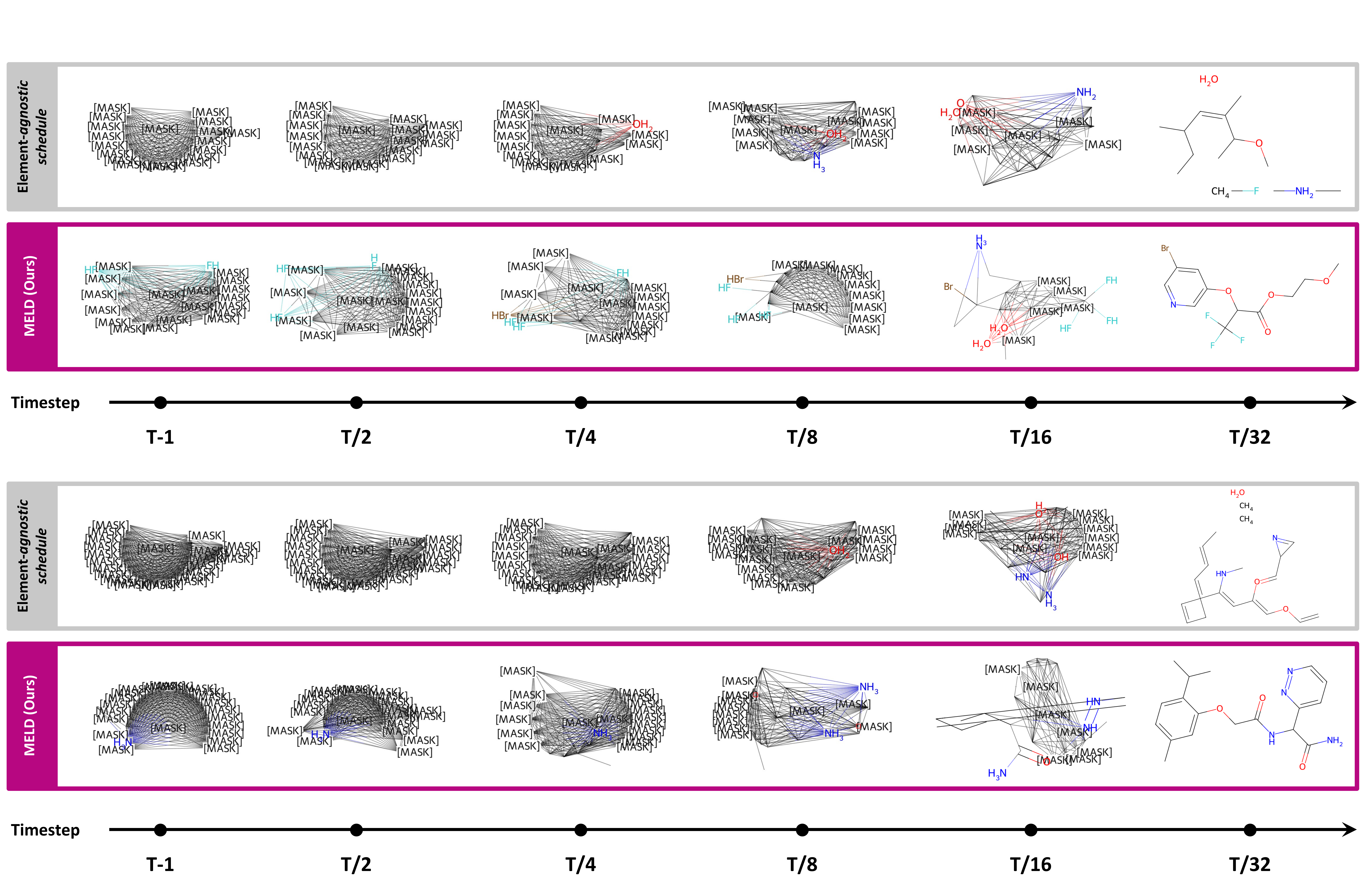}
    % \vspace{-0.2in}
    \caption{
    More comparisons between element-agnostic power-law scheduling and ~\system{} during reconstruction on the ZINC250K dataset. With our proposed noise schedule, most reconstruction occurs at earlier timesteps relative to element-agnostic approach.
    }
    % \vspace{-0.15in}
    \label{fig:appendix_reverse_process}
    \vskip -10pt
\end{figure}

We provide additional visualizations of reverse diffusion trajectories under \system{} compared with those from a fixed power-law (element-agnostic) schedule in~\cref{fig:appendix_reverse_process}. Consistent with our earlier analysis, \system{} achieves faster reconstruction than standard MDMs. For instance, fragments begin to unmask as early as $t = T-1$, whereas the element-agnostic schedule only starts to recover them at $T/4 \leq t \leq T/8$.

\subsection{Molecule visualization}\label{D:mol_vis}
In this section, we provide 2D visualization of molecules generated by \system{}. As illustrated, \system{} generates chemically realistic molecules even for polymers dataset with larger number of atoms (\ie, $|\gV|\le50$), verifying its robustness under various graph sizes.
\begin{figure}[ht]
    \centering
    \includegraphics[width=\linewidth]{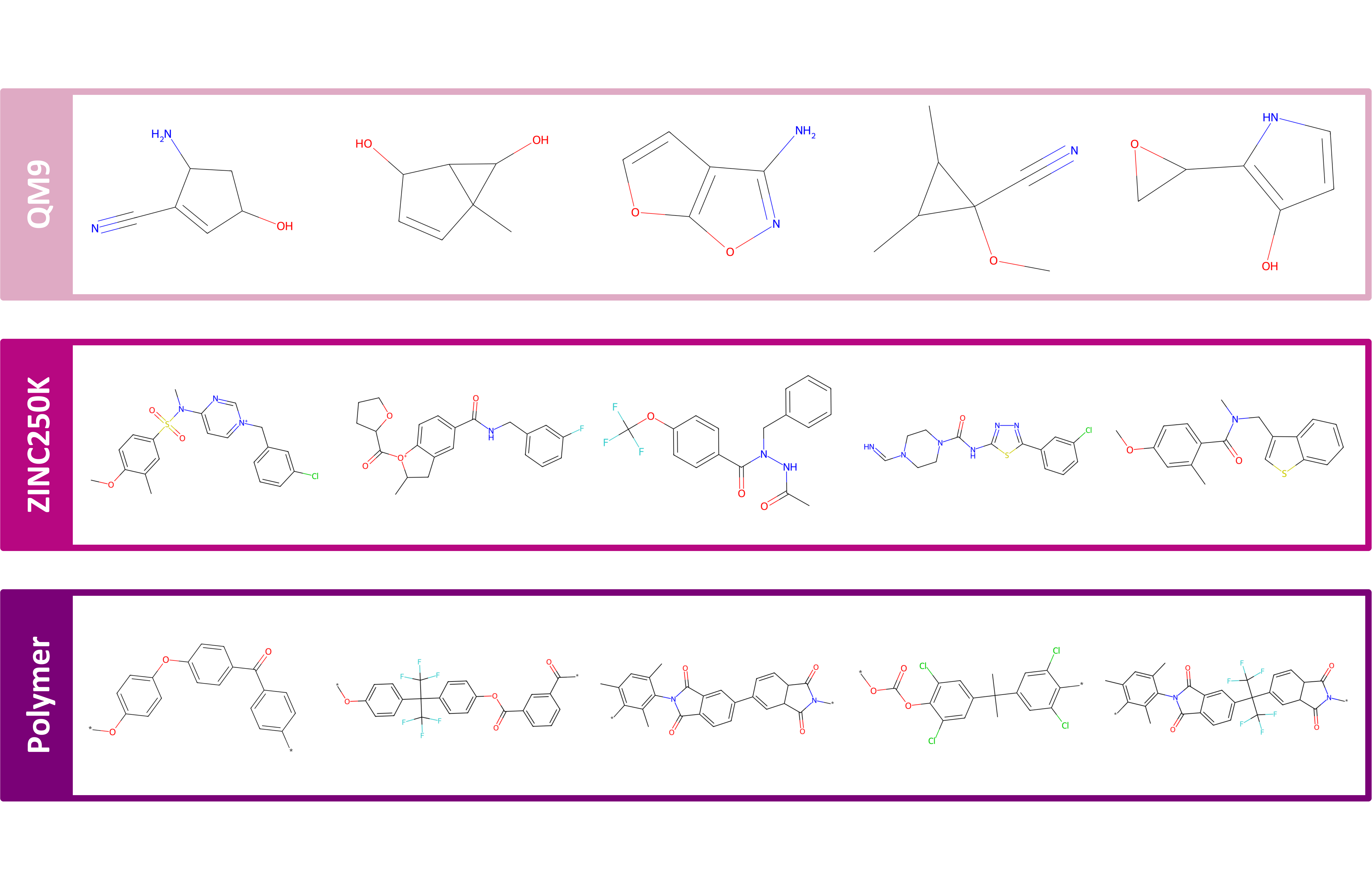}
    % \vspace{-0.2in}
    \caption{
    Visualization of molecules generated by \system{}.
    }
    % \vspace{-0.15in}
    \label{fig:appendix_visualization}
\end{figure}

\end{document}